\documentclass{bmvc2k}

\usepackage{soul}

\usepackage{multirow}
\usepackage{threeparttable}

\title{Median Pixel Difference Convolutional Network for Robust Face Recognition}

\addauthor{Jiehua Zhang}{jiehua.zhang@oulu.fi}{1}
\addauthor{Zhuo Su}{zhuo.su@oulu.fi}{1}
\addauthor{Li Liu\footnotemark[1]\textsuperscript{,2,}}{li.liu@oulu.fi}{1}

\addinstitution{
Center for Machine Vision and Signal
Analysis\\
 University of Oulu\\
 Finland
}
\addinstitution{
 College of System Engineering\\
 National University of Defense
Technology\\
 China
}

\runninghead{Zhang, Su, Liu}{MeDiNet for Robust Face Recognition}

\def\etal{\emph{et al}\bmvaOneDot}

\begin{document}

\maketitle
\footnotetext[1]{\, indicates the corresponding author.}
\begin{abstract}
 Face recognition is one of the most active tasks in computer vision and has been widely used in the real world. With great advances made in convolutional neural networks (CNN), lots of face recognition algorithms have achieved high accuracy on various face datasets. However, existing face recognition algorithms based on CNNs are vulnerable to noise. Noise corrupted image patterns could lead to false activations, significantly decreasing face recognition accuracy in noisy situations. To equip CNNs with built-in robustness to noise of different levels, we proposed a Median Pixel Difference Convolutional Network (MeDiNet) by replacing some traditional convolutional layers with the proposed novel Median Pixel Difference Convolutional Layer (MeDiConv) layer. The proposed MeDiNet integrates the idea of traditional multiscale median filtering with deep CNNs. The MeDiNet is tested on the four face datasets (LFW, CA-LFW, CP-LFW, and YTF) with versatile settings on blur kernels, noise intensities, scales, and JPEG quality factors. Extensive experiments show that our MeDiNet can effectively remove noisy pixels in the feature map and suppress the negative impact of noise, leading to achieving limited accuracy loss under these practical noises compared with the standard CNN under clean conditions. 
\end{abstract}

\section{Introduction}
\label{sec:intro}
Face Recognition is one of the most important research fields in computer vision and pattern recognition. Recent advances in deep learning, coupled with abundant face data, have led to excellent progress in face recognition algorithms \cite{Guo2019,liu2017sphereface,schroff2015face,sun2014face2,sun2014face,wen2014center,deepface2014}. Due to these achievements, face recognition technology is widely utilized in the real world, such as human-computer interaction\cite{lin2019mobile}, video surveillance\cite{chen2019vedio}, and identification \cite{kamgar2011surve,vezzani2013reid}. The accuracy is a crucial metric to evaluate the effectiveness of recognition models. However, the existing noise produced by the sensor of a scanner, cameras, image compression, \textit{etc.}, results in a significant decrease in the accuracy of models. Enhancing the noise-robustness is meaningful to the practical applications of face recognition.

Recently, many studies have been carried out to deal with the challenges of face recognition, such as the variations in expressions, lighting conditions, and poses \cite{cao2018pose,lahasan2019challenge,li2018anti}, with few efforts taking noise into account. The existing noise-robust face recognition methods mainly consider the noisy labels in the face dataset \cite{han2018coteach,wei2019noise,veit2017face,wang2018devil,zhang2020gcn}. This kind of method is hard to adapt to the diversity and complexity of noise in the real world. Our purpose is to improve the robustness of face detectors to noise in the real world.

In this paper, we propose an effective approach to enhance the noise-robustness of CNNs with limited decreasing accuracy compared with standard CNNs. The existing CNNs exhibit sensitivity to noise, particularly salt-and-pepper noise. Inspired by \cite{liu2016med}, we proposed a robust Median Pixel Difference Convolutional Network (MeDiNet), which can effectively suppress noise and improve the performance under noisy conditions by median pixel difference convolution (MeDiConv). Specifically, MRELBP replaces the individual pixel intensities in the feature maps with median representations in the corresponding local regions and reports excellent results in the texture classification task under various noises, showing that combining local median representation is powerful to extract features. The proposed MeDiConv inhabits noise-robust property of MRELBP.  

We summarize our contributions as follows: (\romannumeral1) We proposed a noise-robust network MeDiNet integrating a novel convolutional operation named as MeDiConv to utilize median representation to suppress noise in images effectively. (\romannumeral2) We adopt the image degradation model with a broad range of noises to approximate the noisy model in the real world (blur kernel, noise intensity, scaling, JPEG quality), which can evaluate the performance of our method in practical application. (\romannumeral3) The model can be robust to a variety of noises without generating additional noisy images for training the model. 

\begin{figure}[ht]
  \centering
  \includegraphics[scale=0.35]{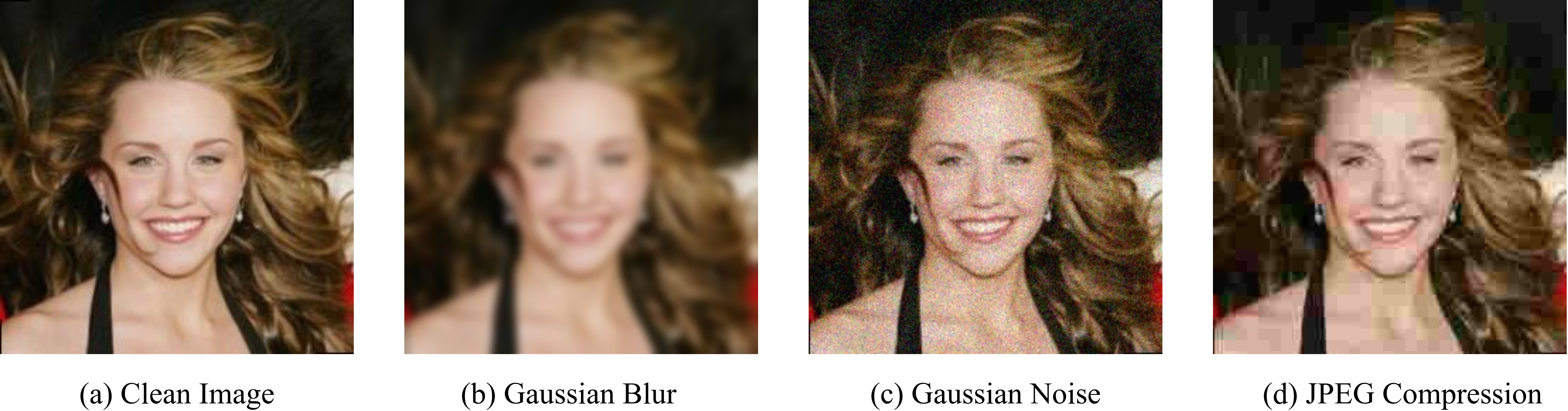}
  \caption{The image example with versatile settings. (b) Gaussian blur, $\sigma$=3, kernel size=13; (c) Gaussian noise, $\sigma$=20; (d) JPEG Compressor, JPEG quality=10.}
  \label{Fig.1}
\end{figure}

In the experiment, MeDiNet is tested on four face datasets with versatile settings. For approximating the complex noise model in the real world, we also add two complex noises to evaluate our method except salt-and-pepper noise and additive white Gaussian noise (AWGN): Heteroscedastic Gaussian (HG) \cite{plotz2017benchmark} and multivariate Gaussian (MG) noise \cite{zhang2018ffd}). The noisy image examples are shown in Figure \ref{Fig.1}. Our experimental results show that the MeDiConv can effectively suppress the negative impact of the noise on the images. We demonstrate that MeDiNet reports superior performance to standard CNN architectures on four face datasets under noisy conditions (LFW \cite{Huang2007}, CA-LFW \cite{Zheng2017}, CP-LFW \cite{zhang2018cpface}, Youtube Face \cite{wolf2011ytf}) while the loss of accuracy on the clean images is limited. 

\section{Related work} 
\subsection{Face recognition in noisy conditions}
With the development of deep learning, a large number of face recognition algorithms have been proposed to address problems like lighting, age, pose \cite{cao2018pose,zhang2018cpface,Zheng2017}. Noise in human face images can significantly decrease the performance of face recognition systems. Noise drops the valuable information in the image and influences the ability of some algorithms to recognize a correct object. To suppress the impact of noise, Ding \etal \cite{deng2017nr} proposed the NR-Network for face recognition under noise, which utilized a multi-input structure to extract a multi-scale and more discriminative feature from the input image. Wu \etal \cite{wu2020} proposed a denoising network D-BSN that is designed as a two-stage training framework with self-supervised learning and knowledge distillation to learn the denoising from unpaired clean and noisy images. Anwar \etal \cite{anwar2019attention} firstly incorporate an attention mechanism to exploit channel dependencies in the denoising task. Li \etal \cite{li2018noise} proposed GFRNet consisting of warping and a reconstruction network. The GFRNet combined the degraded and high-quality images to enhance the performance of face restoration. The existing studies for suppressing noise focus on the restoration tasks \cite{abde2018camera, anwar2019attention,grigrious2017deblur,li2018noise,wu2020,yu2017face}. These methods depend on training with noisy and clean images. We aim at designing a general framework without extra noisy image data to suppress noise in the face recognition task directly. 
\subsection{LBP and convolution}
LBP \cite{ojala2002lbp} was proposed for texture analysis and has been successfully applied to many tasks \cite{texture2018}, such as image retrieval \cite{san2015retr}, object detection \cite{detection2020} and face image analysis \cite{huang2011lbp,bird2019}. LBP has emerged as one of the most prominent texture descriptors. Recently, some studies tried to combine LBP with standard CNNs \cite{pixel2021,yu2020}. The LBCNN was motivated by LBP and utilized a set of fixed sparse pre-defined binary convolutional filters to achieve computational savings \cite{felix2017lbcnn}. Inspired by LBCNN, the LBVCNN added the LBV layer to reduce trainable parameters in facial expression task \cite{kum2019lbvcnn}. Li \etal \cite{li2020att} combined the LBP and attention mechanism to extract useful features, which achieved higher performance in face recognition. These studies demonstrated that combining LBP with standard CNNs is a practical and flexible approach in various tasks. The MRELBP is based on combining a median filter with LBP operation in a multiscale fashion, which can effectively extract noise-resistant patterns. The development of MeDiNet shares similar motivations but has slightly different implementation due to the different processing procedures of LBP and CNN. 

\section{MeDiNet}

The MeDiNet is built on the Sphere20 \cite{liu2017sphereface}. The initial convolutional layers of Sphere20 are replaced with MeDiConv for suppressing the negative impact of a broad range of noises in the face images. The operation of MeDiConv is introduced as follows. 

\subsection{Preliminary of MRELBP}

\begin{figure}[ht]
  \centering
  \includegraphics[scale=0.2]{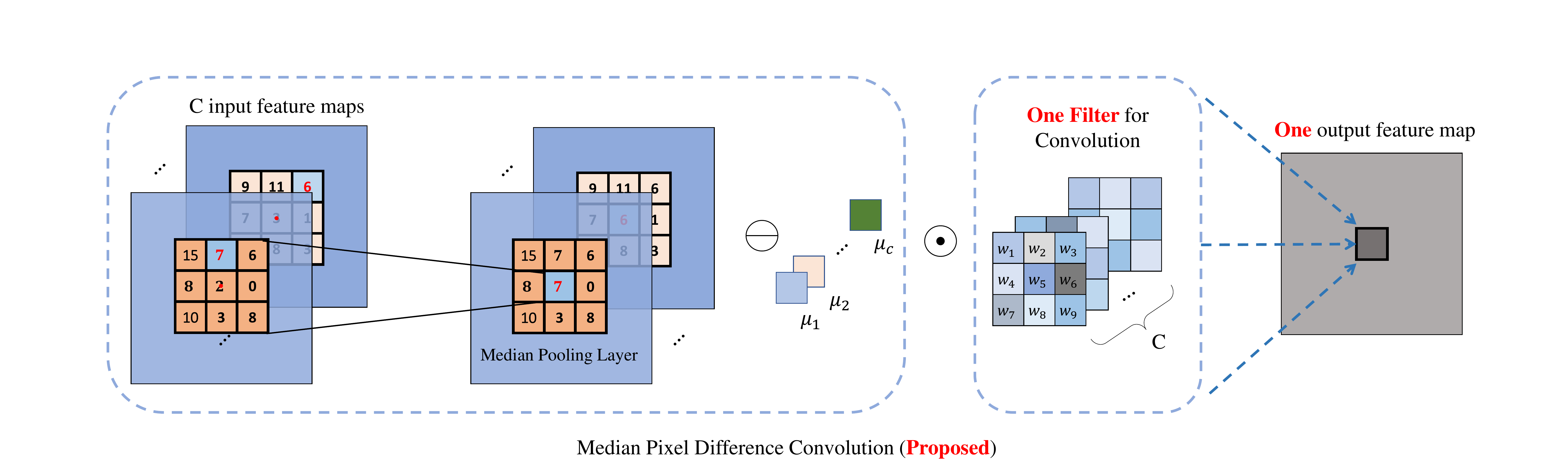}
  \caption{The convolutional operation of Median Pixel Difference Convolution. $\mu_{i}$ denotes a global predefined value of each channel.}
  \label{Fig3}
\end{figure}
The traditional LBP calculates the differences between the pixel value of the central point and those of its neighbors. The MRELBP~\cite{liu2016med} set the pixel intensity as median representation in a local patch, which is more robust to gray scale variations, rotation changes, and noise. We choose the MRELBP descriptor under center pixel representation, a simple and effective way to handle noise. This descriptor can be expressed as:
\begin{equation}
   MRELBP\_CI(x_{c})=s(\phi (X_{c,w})-\mu _{w})
\end{equation}
where $s$ is the sign function, $\phi(X_{c,w})$ denotes the median operation in $X_{c,w}$ (the local patch of size $w\times w$ centered at center pixel $x_{c}$), $\mu$ denotes the mean value of $\phi(X_{i, w})$ over the whole image $X_i$. Inspired by this work, we use a set of learnable weight parameters to replace sign function for combining the standard CNN with median representation. The MRELBP is based on combining a median filter with traditional LBP operation, which can effectively extract powerful features under various noisy conditions. Our MeDiConv also utilize median representation in convolutional operation. This processing can make convolutional layer suppress noisy pixels in forward propagation.

\subsection{MeDiConv}

The calculation process of standard CNN can be described as: the convolutional kernel $W^{c_{out}\times c_{in}\times k\times k}$ slides on the feature map $X^{c_{in}\times H\times W}$ to calculate the output. The formulation can be expressed as following ($c_{out}$ is set to 1 without loss of generality).
\begin{equation}
\label{eq: vanilla}
    X_{out}(u,v)=\sum_{i=0}^{c_{in}-1}\sum_{(p,q)\in \mathcal{N}}^{}X_{i}(u+p,v+q)\cdot W_{i}(p,q) 
\end{equation}
where $(u,v)$ denotes the coordinate of the pixel in feature map, $W_{i}$ denotes the convolutional kernel, and $\mathcal{N}$ denotes the $k\times k$ neighborhood area centered at $(u,v)$. The standard CNNs are sensitive to noise (especially salt-and-pepper noise) since the noise in the local region can confuse the subsequent feature representation process and lead to the detection of false features with no physical correlation.

To suppress noise in regions of feature maps, MeDiConv utilizes local median pixel difference value to replace the original value of $x_{u,v}$ in the convolutional operation. As shown in Figure \ref{Fig3}, the value of each pixel $x_{i}$ in the input feature patch is replaced by the median value of the 3x3 neighborhood centered on $x_{i}$. The procedure can be decomposed as the following stages: 1) for each location $(u, v)$ with its neighborhood region $\{(u+p, v+q) | (p,q)\in \mathcal{N}\}$, calculating the median representation for each pixel located in the neighborhood region; 2) replacing the pixel intensity $X_i(u+p, v+q)$ in Eq.~\ref{eq: vanilla} with the difference between the calculated median representation of $X_i(u+p, v+q)$ and a global predefined value $\mu$; 3) combining the output in stage 2 with the weights to generate the output feature map. Therefore, MeDiConv can be expressed as:

\begin{equation}
   \tilde{X}_{out}(u,v)=\sum_{i=0}^{c_{in}-1}\sum_{(p,q)\in \mathcal{N}}^{}(\phi(X_{i}(u+p,v+q))-\mu )\cdot W_{i}(p,q)
\end{equation}


\begin{equation}
    \phi(X_i(u,v))=Median(\{X_i(u+p,v+q) | (p,q)\in \mathcal{N}\})
\end{equation}

where $\tilde{X}_{out}$ is the output feature map, $\phi(X_i(u,v))$ denotes the median representation of the $k\times k$ region centered at $X_i(u,v)$, and $\mu$ represents the average value of $\phi(X_i(u,v))$ over the whole feature map. The aim of subtracting $\mu$ is to eliminate the bias of pixel intensities in the feature map. 

 Compared with standard convolution, the MeDiConv is a nonlinear smoothing operation, which can effectively remove outliers with limited impact on the ability of feature extraction \cite{median1981}. Noise corrupted image patterns could lead to false activations, leading to a significant decrease in accuracy. For MeDiConv, the effect of applying MeDiConv at multiple layers can be considered as applying multiple median filters of different kernel sizes on the original image as each MeDiConv layer has a different receptive field. That is also why the proposed MeDiNet can deal with noise of different levels.  Considering the trade-off between accuracy and noise robustness, MeDiNet only set MeDiConv in the first few layers. When the standard CNN in the deep layer of the network is replaced by MeDiConv, the response of the feature map will be degraded. Setting MeDiConv in the beginning layers is sufficient to eliminate the impact of noise.

\section{Experiments}
\subsection{Dataset}
We adopt the CASIA-WebFace \cite{dong2014casia} as a training dataset. The LFW \cite{Huang2007}, CA-LFW \cite{Zheng2017}, CP-LFW \cite{zhang2018cpface}, Youtube Face \cite{wolf2011ytf} are used for evaluation. The details of datasets are as following:

\textbf{CASIA-WebFace}. The CASIA-WebFace dataset is collected from IMDb website containing 10,575 subjects and 494,141 images. The collection and annotation of CASIA-WebFace use a semi-automatical way. It also utilizes the name bag and face similarity for efficient labeling.

\textbf{LFW}. Labeled Faces in the Wild (LFW) aims to study face recognition in an unconstrained environment, containing more than 13,000 face images from the website. Each face in the LFW is marked with a person's name, and about 1680 people contain more than two faces. Although there are some mislabeled matched pairs, LFW is considered completely clean. 

\textbf{CA-LFW and CP-LFW}. CA-LFW and CP-LFW are renovations of LFW dataset. These datasets focus on the cross-age and cross-pose face challenges in face recognition. The images are collected from 3,000 positive face pairs with age gap and pose gap, respectively. Compared with the LFW, both datasets are more challenging with lower test accuracy. 

\textbf{YouTube Faces}. The YouTube Faces Dataset (YTF) is a facial video database used to study unconstrained face recognition in videos. The dataset contains 3,425 videos from 1,595 different people. All the videos are collected from YouTube. Each subject provides an average of 2.15 videos. The shortest clip length is 48 frames, the longest clip length is 6070, and the average video clip length is 181.3.

\subsection{Degradation model}
\label{section:3.2}
In the practical application, the quality of images can be affected by noise, scaling, compression, and their combination. For verifying the robustness of MeDiNet, a degradation model is required to generate realistic noisy images \cite{li2018noise}, which can be defined as:
\begin{equation}
   I^{d}=((I\otimes k )_{\downarrow_{S}}+n)_{JPEG_{C}}
\end{equation}
where $\otimes$ means the convolutional operation. $k$ denotes the blur kernel. $\downarrow_{S}$ denotes the downsampling operation with the scale factor $S$. $n$ denotes the noise like AWGN. $(\cdot )_{JPEG_{\mathcal{C}}}$ denotes the JPEG compression with the quality factor $c$. The degradation $ I^{d}=((I\otimes k )_{\downarrow_{S}}+n)$ is caused by remote acquisition and  $(\cdot )_{JPEG_{\mathcal{C}}}$ expresses the effect of JPEG compression.

The parameter settings of our degradation model is shown as below:

\textbf{Blur kernel.} In our paper, we utilize Gaussian blur kernel and Motion blur kernel as the Simulation of defocusing effect. The variance $\varrho$  of Gaussian blur is sampled from the set  $\left \{1:1:3  \right \}$. The degree $\mathcal{D}$ of Motion blur is sampled from the set $\left \{10:10:40  \right \}$. 0 means no blur operation.

\textbf{Downsampler.} The scale factor $\mathcal{S}$ is sampled from the set $\left \{2,4,6,8  \right \}$. It should be noted that the image will be upsampled to its original size after downsampling.

\textbf{Noise.} Besides common noise (like AWGN and salt-and-pepper noise), we select more complex noise to evaluate our model, including HG $n_{i}\sim \mathcal{N}(0,\alpha ^{2}x_{i}+\delta^{2})$ and MG noise $n\sim \mathcal{N}(0,\sum)$ with $\sum= \mathcal{L}^{2}\cdot U\Lambda U^{T}$. $U$ is a random unitary matrix, $\Lambda$ is a diagonal matrix of three random values in the range (0,1). The sampling ranges of noisy parameters are following: 1) AWGN: $\sigma$ in $\left \{15,25,50  \right \}$; 2) Salt-and-pepper noise: $\rho$ in $\left \{5\%,10\%,15\%,25\%  \right \}$; 3) HG noise: $\alpha$ in $\left \{20:10:40  \right \}$ , $\sigma$ in $\left \{10,20  \right \}$; 4) MG noise: $\mathcal{L}=75$. 

\textbf{JPEG compression.} When storing images, the quality would be compressed. We add the JPEG Compression with the quality factor $\mathcal{C}$ on the degraded images. The $\mathcal{C}$ is sampled from $\left \{10:10:40  \right \}$. 0 means the image is losslessly compressed.

\subsection{Experimental settings}
Our model is trained on the CASIA-WebFace dataset. We crop the training image with size 112x96 to feed the network. The model is trained in 20 epochs with SGD optimizer. The batchsize is set as 256. The learning rate is initialized as 1e-2 and decayed by factor 0.1 after 10, 15, 18 epoch. We add the degradation model on test images for evaluation and conduct a separate experiment to analyze the influence of independent noise.  

\subsection{Experiments on the images under single noise}
\label{section:4.3}
\begin{figure}[ht]
  \centering
  \includegraphics[width=0.95\linewidth]{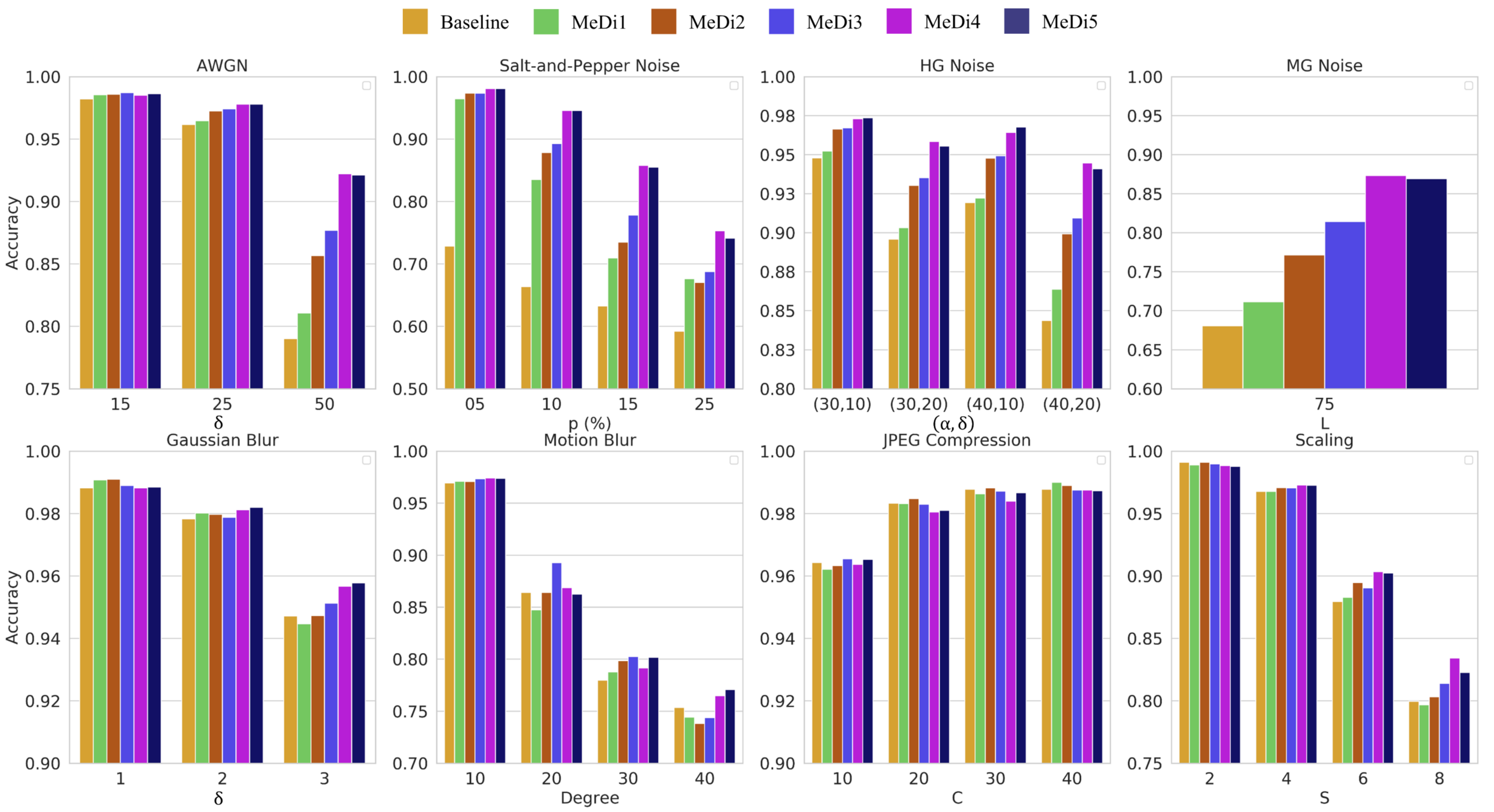}
  \caption{The evaluation on the LFW dataset under single noise.}
  \label{Fig4}
\end{figure}

\begin{figure}[ht]
  \centering
  \includegraphics[width=0.95\linewidth]{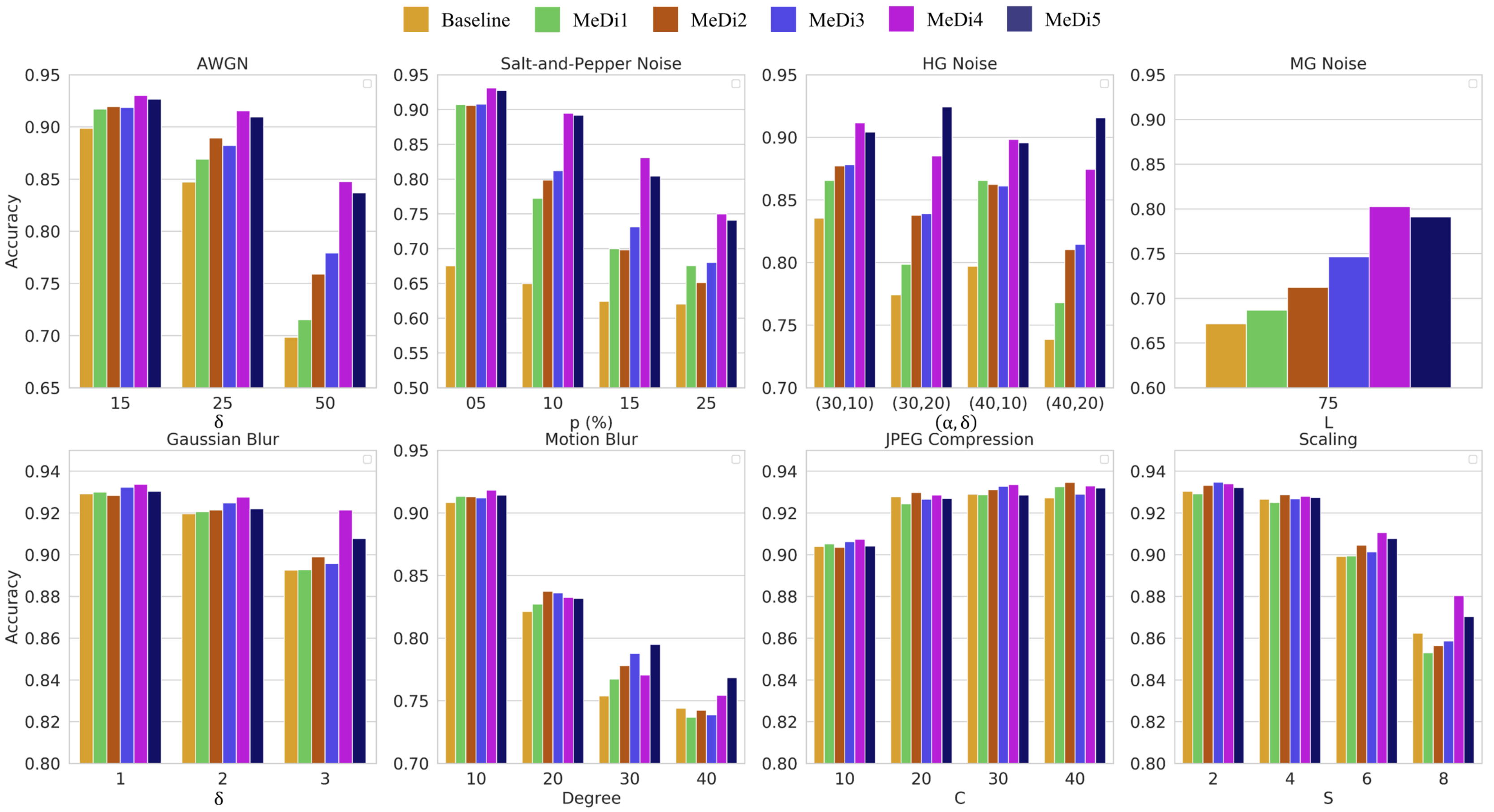}
  \caption{The evaluation on the YTF dataset under single noise.}
  \label{Fig5}
\end{figure}

Our main objective of this work is to enhance the inherent noise robustness of CNNs without significant performance loss. Therefore, Our method focuses on applying median representation to the CNN. We set the Sphere20 as baseline \cite{liu2017sphereface} and compare the performance of different MeDiConv layers. The MeDiNet is built on the Sphere20. The first to fifth layers of Sphere20 are replaced with the MeDiConv (For example, the MeDi3 denotes the first three layers are MeDiConv). The training stage only used clean images. The noisy images were used in the test stage. Various types of noise are utilized to evaluate our training model on the LFW and YTF datasets (See Figure ~\ref{Fig4} and Figure~\ref{Fig5}). We separately analyzed the robustness of MeDiNet to each noise type. \emph{The detailed information of results can be observed in the supplemental material}. 

\emph{Evaluation on blurring}. In practical applications, the blurred images are generated from defocus and pre-processing stage. We evaluate MeDiNet on the gaussian and motion blur kernel. The blurring leads to the degradation of edges and structures in the image. The results illustrate that our method is robust to image blurring. However, the performance drops slightly with the number of MeDiConv increasing to five layers.

\emph{Evaluation on AWGN and Salt-and-Pepper noise}. These noises are generated from remote sensing and can be regarded as outliers in the images. The standard CNNs are sensitive to noise. Noise pixels can significantly reduce the accuracy of face recognition, especially salt-and-pepper noise. When setting the proportion of salt-and-pepper noise to 20\%, the accuracy of model drops by more than 30\%. MeDiNet can effectively remove the outliers in the feature maps by median representation. The test results in Figure ~\ref{Fig4} and Figure~\ref{Fig5} show that our method can enhance the robustness to a broad range of noises.

\emph{Evaluation on complex noise}. However, CNN usually generates poorly due to complex noise even if it can adapt to AWGN and salt-pepper noise \cite{plotz2017benchmark}. Thus, we also present the results on complex noise models: HG and MG noise (See Figure ~\ref{Fig4} and Figure~\ref{Fig5}). We can observe that the improvement is still significant, which can prove the robustness of proposed MeDiNet.

\emph{Evaluation on Scaling and Compression}. Standard CNNs achieve slight accuracy loss under compression. In the procedure of JPEG compression, part of high-frequency information is removed, which has a negative impact on the ability of CNN to extract features. For scaling, results demonstrate that the accuracy drops significantly when enlarging scaling factor $\mathcal{S}$ because of sampling and interpolation. The detailed information of images might be removed during the scaling process. The median representation can effectively enhance the expression ability of features to a certain extent. 

\vspace{-10pt}
Through the analysis of each experimental result under the single noise, we can observe that the MeDiNet inherits the characteristics of MRELBP, and demonstrates striking robustness to various image noises, including blurring, AWGN, Salt-and-Pepper, HG, MG, Compression and Scaling. In the Sections \ref{section:4.4}, the combination of multiple noises is adopted to test our method, which approximates noise interference in the real world. 
\vspace{-5pt}
\begin{figure}[ht]
  \centering
  \includegraphics[width=0.8\linewidth]{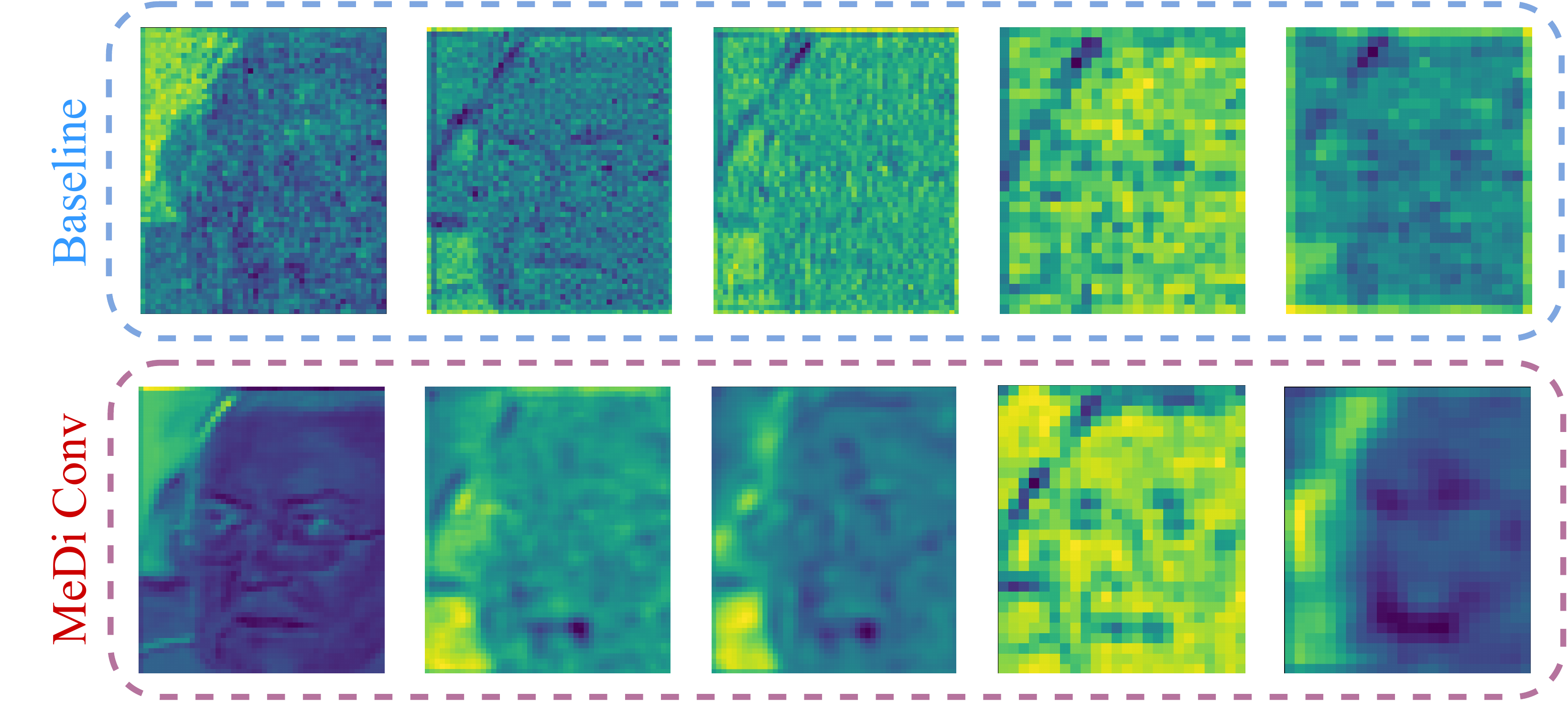}
  \caption{The visualization of feature map. The first row are outputs of the first to fifth layer in the baseline; the second row are outputs of the last MeDiConv layer in the MeDi1-5 (\textit{e.g.,} the second figure denote the feature map of last MeDiConv layer in MeDi2).}
  \label{Fig6}
\end{figure}
\vspace{-8pt}
We visualized feature maps of middle layers in the model to demonstrate the ability of MeDiConv to suppress noise (See Figure \ref{Fig6}). The input image was added with the AWGN with $\sigma=50$.  We can observe that the output by MeDiConv is smooth and contains fewer noisy pixels. The structure and edges of face are identifiable, leading to improve the accuracy of face recognition.

\subsection{Experiments on the degraded images}
\label{section:4.4}
In this section, the degradation model introduced in Sections \ref{section:3.2} is adopted to approximate the real-world noise. We utilize two blurring methods and four types of noise to combine. The parameters of the degradation model are randomly generated within their respective parameter ranges in Sections \ref{section:3.2} (\emph{The value of parameters can be observed in the supplemental material}). In this section, we present the evaluation of four combining noises: 

$\bullet$ Gaussian$\_$blur+AWGN+Scaling+Compression (GB+AWGN);

$\bullet$ Gaussian$\_$blur+Salt-and-Pepper+Scaling+Compression (GB+SP);

$\bullet$ Gaussian$\_$blur+HG+Scaling+Compression (GB+HG);

$\bullet$ Gaussian$\_$blur+MG+Scaling+Compression (GB+MG). 

The \emph{clean} denotes no noise in the images. Under real-world noisy conditions, the face recognition accuracy drops more significantly than it under single noise. Tab.\ref{tab:Tab1} and Tab.\ref{tab:Tab2} show the evaluation of four datasets under real-world noisy conditions. \emph{The results of other combinations can be observed in the supplemental material.} 

\begin{table}[ht]
\centering
\caption{The evaluation on the LFW and YTF dataset}
\begin{threeparttable}
\label{tab:Tab1}
\setlength{\tabcolsep}{0.02\linewidth}
\resizebox*{0.99\linewidth}{!}{
\begin{tabular}{|c|c|c|c|c|c|c|c|c|c|c|}
\hline
               & \multicolumn{2}{c|}{Clean}        & \multicolumn{2}{c|}{GB+AWGN}      & \multicolumn{2}{c|}{GB+SP}        & \multicolumn{2}{c|}{GB+HG}        & \multicolumn{2}{c|}{GB+MG}        \\ \hline
               & LFW             & YTF             & LFW             & YTF             & LFW             & YTF             & LFW             & YTF             & LFW             & YTF             \\ \hline
Sphere20       & \textbf{0.9922} & 0.9330          & 0.7682          & 0.7636          & 0.6762          & 0.6618          & 0.7482          & 0.7440          & 0.6573          & 0.6628          \\ \hline
MeDi1 & 0.9908          & 0.9300          & 0.7938          & 0.7946          & 0.7407          & 0.7472          & 0.7790          & 0.7500          & 0.6723          & 0.6790          \\ \hline
MeDi2 & 0.9910          & 0.9322          & 0.8390          & 0.8256          & 0.7797          & 0.7848          & 0.8192          & 0.7984          & 0.6755          & 0.6754          \\ \hline
MeDi3 & 0.9898          & 0.9328          & 0.8402          & 0.8254          & 0.7908          & 0.7888          & 0.8208          & 0.8058          & \textbf{0.7265} & 0.7178          \\ \hline
MeDi4 & 0.9882          & \textbf{0.9350} & \textbf{0.9083} & \textbf{0.8864} & \textbf{0.8820} & \textbf{0.8782} & \textbf{0.8942} & \textbf{0.8708} & 0.7157          & \textbf{0.7654} \\ \hline
MeDi5 & 0.9892          & 0.9334          & 0.8943          & 0.8812          & 0.8735          & 0.8648          & 0.8848          & 0.8590          & 0.7222          & 0.7452          \\ \hline
\end{tabular}
}
\begin{tablenotes}
    \footnotesize
    \item[*] The MeDi1 can be regarded as the image preprocessed by median filtering.   
    \end{tablenotes}
\end{threeparttable}
\end{table}
\vspace{-6pt}
Compared with the single noise in Section \ref{section:4.3}, the degradation model further increases the complexity of noise. The results in Tab. \ref{tab:Tab1} and Tab. \ref{tab:Tab2} demonstrate that MeDiNet can enhance the performance under real-world noisy condition. We can observe improvements of GB+AWGN, GB+SP, and GB+HG are significant. The performance under the GB+MG is still limited. 

In order to fully verify the effectiveness of our method, we also report the results of CA-LFW and CP-LFW dataset (See Tab. \ref{tab:Tab2}). These datasets are collected for specific challenges (pose and age difference). We can observe that the improvement on the CP-LFW dataset is limited. This dataset selects face pairs with pose differences to add pose variation to intra-class variance and contains less edge information than LFW. When using the MeDiConv, the degradation of edge leads to a limited increase on the CP-LFW (e.g., the accuracy of MeDi5 is only 3.32\% higher than baseline under GB+MG condition).  

\begin{table}[ht]
\centering
\caption{The evaluation on the CA-LFW and CP-LFW dataset}
\label{tab:Tab2}
\setlength{\tabcolsep}{0.025\linewidth}
\resizebox*{0.99\linewidth}{!}{
\begin{tabular}{|c|c|c|c|c|c|c|c|c|c|c|}
\hline
               & \multicolumn{2}{c|}{Clean}        & \multicolumn{2}{c|}{GB+AWGN}      & \multicolumn{2}{c|}{GB+SP}        & \multicolumn{2}{c|}{GB+HG}        & \multicolumn{2}{c|}{GB+MG}        \\ \hline
               & CP              & CA              & CP              & CA              & CP              & CA              & CP              & CA              & CP              & CA              \\ \hline
Sphere20       & 0.7618          & \textbf{0.9078} & 0.5697          & 0.6957          & 0.5238          & 0.5545          & 0.5502          & 0.6030          & 0.5130          & 0.5403          \\ \hline
MeDi1 & 0.7607          & 0.9020          & 0.5868          & 0.7288          & 0.5525          & 0.6500          & 0.5500          & 0.6405          & 0.5133          & 0.5337          \\ \hline
MeDi2 & \textbf{0.7638} & 0.9032          & 0.5927          & 0.7647          & 0.5677          & 0.6903          & 0.5675          & 0.6835          & 0.5315          & 0.5498          \\ \hline
MeDi3 & 0.7620          & 0.9017          & 0.6028          & 0.7632          & 0.5660          & 0.6940          & 0.5707          & 0.6912          & 0.5405          & 0.5577          \\ \hline
MeDi4 & 0.7407          & 0.8938          & \textbf{0.6485} & \textbf{0.8097} & \textbf{0.6210} & 0.7560          & \textbf{0.6097} & 0.7558          & 0.5372          & 0.6322          \\ \hline
MeDi5 & 0.7553          & 0.8927          & 0.6438          & 0.8082          & 0.6132          & \textbf{0.7628} & 0.6028          & \textbf{0.7582} & \textbf{0.5462} & \textbf{0.6327} \\ \hline
\end{tabular}
}
\end{table}
\vspace{-6pt}
The evaluation of four datasets demonstrates that proposed MeDiNet enhances the performance of face recognition under the real-world noisy conditions. The accuracy is improved when the number of MeDiConv layers increases. The median representation causes a probability of missing edge information. Thus, the performance slightly drops compared with baseline when testing on clean images. When adding multiple combinations of noises, the face recognition accuracy increases significantly on four datasets compared with baseline due to the noise-robustness of MeDiNet. However, the improvement under complex noise (GB+HG and GB+MG) is lower than other combinations. The ability of MeDiNet to deal with complex noises needs to be strengthened. Solving the problem of edge caused by median representation is also the focus of future work.  

In order to further analyze the characteristics of MeDiConv, we report the time consumption of the model and transfer it to the ResNet architecture for evaluation.

\begin{figure}[!ht]
  \centering
  \includegraphics[width=1.0\linewidth]{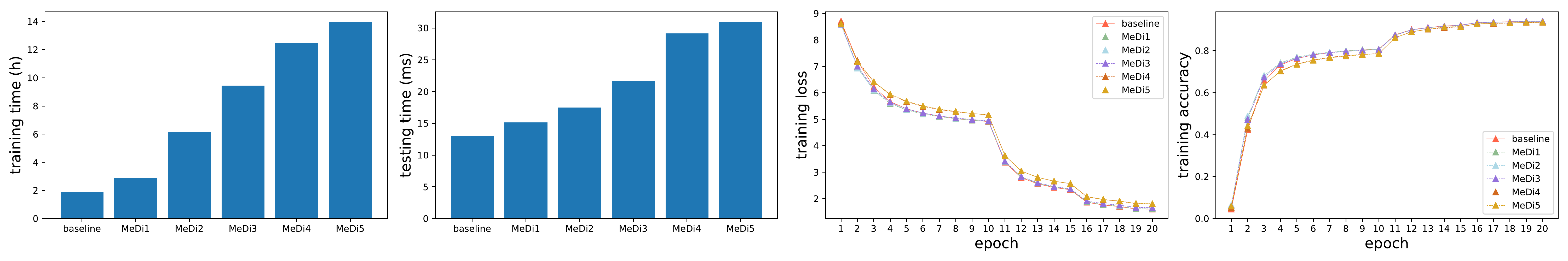}
  \caption{The first graph shows the training time of each model (20 epochs); The second graph shows the inference time for per image; The third and fourth graphs show the training loss and accuracy.}
  \label{Fig7}
\end{figure}

\textbf{Time Consumption} In the experiment, we observed median operation is hard to speed up on GPU. When the number of MeDiConv layers increases, the training time consumption significantly increases (See in Fig. \ref{Fig7}). It would be more efficient to implement from scratch at C/Cuda level. For training convergence, the difference between the loss curve and training accuracy of MeDi1-3 and baseline is limited, illustrating that training MeDiNet is stable.

\vspace{-6pt}
\textbf{Transfer Ability} For verifying that MeDiConv is a general way to handle noise, we choose ResNet18 as the backbone. The result is shown in Tab \ref{tab:Tab3}. We fixed the first convolution in ResNet as MeDiConv. The MeDiRes1 denotes that we replace the convolutional layers in the first block with MeDiConv. The result showed our methods still can suppress the negative impact of noise, which can prove that MeDiConv is a general way to handle noise that can be transferred to other architectures. 
\vspace{-6pt}
\begin{table}[!ht]
\centering
\caption{The ResNet architecture evaluated on the LFW}
\label{tab:Tab3}
\setlength{\tabcolsep}{0.025\linewidth}
\resizebox*{0.7\linewidth}{!}{
\begin{tabular}{|c|c|c|c|c|c|}
\hline
         & Clean         & GB+AWGN       & GB+SP         & GB+HG         & GB+MG         \\ \hline
ResNet18 & \textbf{0.9888} & 0.8065 & 0.7000 & 0.7943 & 0.6665 \\ \hline
MeDiRes1 & 0.9730 & 0.8348 & 0.7847 & 0.8205 & 0.6688 \\ \hline
MeDiRes2 & 0.9810 & \textbf{0.8580} & \textbf{0.8110} & \textbf{0.8378} & \textbf{0.6950} \\ \hline
\end{tabular}
}
\end{table}
\vspace{-20pt}
\section{Conclusion}
We proposed a robust architecture MeDiNet, to enhance the performance of face recognition under noisy conditions. Standard CNNs are sensitive to the noise in the real world, especially in low-quality security cameras, low-light conditions, etc., which leads to a significant decrease in the accuracy of face recognition. Inspired by MRELBP descriptor, MeDiNet adopted median pixel difference convolutions, which can effectively remove noisy pixels in the feature map. Extensive experimental results showed the MeDiNet demonstrates striking robustness to various image noises. 

Compared with standard CNNs, our method has attractive properties on noisy images. However, there are several challenges that need to be addressed: 1) The trade-off exists between accuracy and noisy robustness. The median representation can affect feature extraction (e.g., edge detection) in the deep layers; 2) The overlap region between neighbor MeDiConv kernel might result in the same median value, which can affect the accuracy of face recognition; 3) The accuracy still decreases significantly under the complex noise.  

\setcounter{secnumdepth}{0}
\section{Acknowledgement}
This work was partially supported by the Academy of Finland under grant 331883 and the National Natural Science Foundation of China under Grant 61872379. 
The authors also wish to acknowledge CSC IT Center for Science, Finland, for computational resources.

\bibliography{egbib}
\end{document}